\documentclass[sigplan,10pt,review=false,nonacm=true]{acmart}

\usepackage{graphicx}
\usepackage{subfig}
\usepackage{algorithm}
\usepackage[noend]{algpseudocode}
\usepackage{mathtools}
\usepackage{pgf}
\usepackage{mleftright}
\usepackage{nicefrac}
\usepackage{varwidth}
\usepackage{comment}
\usepackage{multirow}
\usepackage{paralist}
\renewcommand\footnotetextcopyrightpermission[1]{} 

\algnewcommand{\LineComment}[1]{\State \(\triangleright\) #1}
\algnewcommand\And{\textbf{and}}

\algnewcommand\algorithmicforeach{\textbf{for each}}
\algdef{S}[FOR]{ForEach}[1]{\algorithmicforeach\ #1\ \algorithmicdo}

\usepackage{tikz}
\newcommand*\circled[1]{\tikz[baseline=(char.base)]{
            \node[shape=circle,draw,inner sep=1pt] (char) {#1};}}

\AtBeginDocument{%
  \providecommand\BibTeX{{%
    \normalfont B\kern-0.5em{\scshape i\kern-0.25em b}\kern-0.8em\TeX}}}
    
\renewcommand\footnotetextcopyrightpermission[1]{} 

\AtBeginDocument{%
  \providecommand\BibTeX{{%
    \normalfont B\kern-0.5em{\scshape i\kern-0.25em b}\kern-0.8em\TeX}}}

\newif\ifComments
\Commentstrue

\ifComments
\newcommand{\del}[1]{\noindent\textcolor{gray}{{#1}}}

\newcommand{\victor}[1]{\noindent\textcolor{magenta}{Victor: {#1}}}
\newcommand{\rafael}[1]{\noindent\textcolor{orange}{Rafael: {#1}}}
\newcommand{\joao}[1]{\noindent\textcolor{magenta}{Joao: {#1}}}
\newcommand{\nelson}[1]{\noindent\textcolor{green}{Nelson: {#1}}}
\newcommand{\jose}[1]{\noindent\textcolor{magenta}{Jose: {#1}}}
\newcommand{\guido}[1]{\noindent\textcolor{violet}{Guido: {#1}}}
\else
\newcommand{\del}[1]{}

\newcommand{\victor}[1]{}
\newcommand{\rafael}[1]{}
\newcommand{\joao}[1]{}
\newcommand{\nelson}[1]{}
\newcommand{\jose}[1]{}
\newcommand{\guido}[1]{}
\fi

\newcommand{\etal}{{\em et al. }}

\newcommand{\rsec}[1]{Section~\ref{sec:#1}}
\newcommand{\rssec}[1]{Subsection~\ref{sec:#1}}

\newcommand{\rtab}[1]{Table~\ref{tab:#1}}
\newcommand{\rfig}[1]{Figure~\ref{fig:#1}}

\newcommand{\req}[1]{Equation~\ref{eq:#1}}
\newcommand{\reqs}[2]{Equations~\ref{eq:#1} --~\ref{eq:#2}}

\newcommand{\tit}[1]{{\textit{#1}}}

\newcommand{\name}{SConv}
\newcommand{\base}{Base}
\begin{document}

\title{Advancing Direct Convolution using Convolution Slicing Optimization and ISA Extensions}


\author{Victor Ferrari}
\authornote{Both authors contributed equally to this research.}
\email{v187890@dac.unicamp.br}
\affiliation{
    \institution{Institute of Computing - UNICAMP}
    \country{Brazil}
}

\author{Rafael Sousa}
\authornotemark[1]
\email{rafael.sousa@ic.unicamp.br}
\affiliation{
    \institution{Institute of Computing - UNICAMP}
    \country{Brazil}
}

\author{Marcio Pereira}
\email{mpereira@ic.unicamp.br}
\affiliation{
    \institution{Institute of Computing - UNICAMP}
    \country{Brazil}
}

\author{Jo\~ao P. L. de Carvalho}
\email{joao.carvalho@ualberta.ca}
\affiliation{
    \institution{University of Alberta}
    \country{Canada}
}

\author{Jos\'e Nelson Amaral}
\email{jamaral@ualberta.ca}
\affiliation{
    \institution{University of Alberta}
    \country{Canada}
}

\author{Jos\'e Moreira}
\email{jmoreira@us.ibm.com}
\affiliation{
    \institution{IBM Research}
    \country{United States of America}
}

\author{Guido Araujo}
\email{guido@unicamp.br}
\affiliation{
    \institution{Institute of Computing - UNICAMP}
    \country{Brazil}
}

\renewcommand{\shortauthors}{Ferrari and Sousa, et al.}

\begin{abstract}
Convolution is one of the most computationally intensive operations that must be performed for machine-learning model inference. A traditional approach to compute convolutions is known as the Im2Col + BLAS method. This paper proposes \name: a direct-convolution algorithm based on a MLIR/LLVM code-generation toolchain that can be integrated into machine-learning compilers   . This algorithm introduces:
\begin{inparaenum}[ (a)]
\item Convolution Slicing Analysis (CSA) ---  a convolution-specific 3D cache-blocking analysis pass  that focuses on tile reuse over the cache hierarchy;
\item Convolution Slicing Optimization (CSO) --- a code-generation pass that uses CSA to generate a tiled direct-convolution macro-kernel; and
\item Vector-Based Packing (VBP) ---  an architecture-specific optimized input-tensor packing solution based on vector-register shift instructions for convolutions with unitary stride.
\end{inparaenum}
Experiments conducted on 393 convolutions from full  ONNX-MLIR machine-learning models indicate that the elimination of the Im2Col transformation and the use of fast packing routines result in a total packing time reduction, on full model inference, of 2.0x -- 3.9x on Intel x86 and 3.6x -- 7.2x on IBM POWER10. The speed-up over an Im2Col + BLAS method based on current BLAS implementations for end-to-end machine-learning model inference is in the range of 9\% -- 25\% for Intel x86 and 10\% -- 42\% for IBM POWER10 architectures. The total convolution speedup for model inference is 12\% -- 27\% on Intel x86 and 26\% -- 46\% on IBM POWER10. \name~also outperforms BLAS GEMM, when computing pointwise convolutions, in more than 83\% of the 219 tested instances. 
\end{abstract}




\maketitle

\section{Introduction}
\label{sec:introduction}

Convolution is a mathematical tensor operation commonly used for image processing and in \textit{Convolutional Neural Network (CNN)} models. Convolution is computationally intensive and it accounts for most of the execution time of a typical CNN model. Thus, many approaches have been proposed to speed up convolution~\cite{im2col, convgemm, indirect, cudnn, kn2row2, direct_conv}.
The most  well-known  approach for convolution relies on a sequence of two major steps:
\begin{inparaenum}[(a)]
\item Im2Col operation to pack  the input image and filters into  matrices; and
\item GEMM (Generic Matrix Multiplication) to compute the final convolution result.
\end{inparaenum}
Depending on the convolution layer, the matrices resulting from Im2Col can become very large, potentially leading to poor memory-hierarchy performance if GEMM is not well-optimized. To address the issue, this approach relies on efficient cache-optimized implementations of GEMM available from libraries such as Eigen~\cite{eigenweb} and BLAS~\cite{openblas}. These implementations divide the matrices into smaller blocks called tiles that are suitable to the cache hierarchy and then run a machine-dependent GEMM \textit{micro-kernel} on each tiled block.



In previous work direct convolution outperforms the traditional Im2Col followed by GEMM approach under certain conditions \cite{direct_conv, direct_arm}. 
This paper presents \name: a direct-convolution algorithm that uses architectural information to improve convolution's cache utilization and ISA extensions to accelerate data packing and computation, suitable for SIMD architectures. This paper also evaluates \name~in full machine-learning model inference on two architectures (Intel x86 and IBM POWER). The algorithm can leverage Instruction-Set Architecture (ISA) acceleration extensions (\tit{e.g.} IBM POWER10 MMA) to compete with optimization libraries such as BLAS. 


Compared with Im2Col followed by GEMM, \name~reduces the data-manipulation overhead and uses a specific cache-tiling technique to reduce the number of cache misses and improve performance. 
The traditional approach
\begin{inparaenum}[(a)]
\item applies Im2Col to the input image and filter set, potentially creating large matrices;
\item applies cache tiling to the matrices to determine the best tile size and their allocation at each cache level (as in ~\cite{goto}); 
\item changes the data layout of a tile with a packing routine; and 
\item computes GEMM using a micro-kernel.
\end{inparaenum}
In the proposed algorithm, (a) and (c) are replaced by a single packing step executed after tiling. The central ideas of this paper are encapsulated in two new passes in the software stack.
First, a \textit{Convolution Slicing Analysis} (CSA) pass takes the input image, filters, and cache sizes and estimates the best tile sizes, their allocation on the memory hierarchy, and the order in which they should be accessed --  CSA is the convolution equivalent to the Goto \etal~\cite{goto} algorithm for matrix multiplication. 
Second, a new \textit{Convolution Slicing Optimization} (CSO) code-generation pass synthesizes an optimized loop nesting structure that executes the tiled convolution. Finally, this loop structure uses the packing algorithm to compact the input image according to the outer-product-based micro-kernel, which computes the convolution.

After presenting convolution (\rsec{background}), this paper makes the following contributions. 
\begin{compactitem}
   \item A compiler-based solution for convolution code generation (\rsec{overview}) that does not depend on linking with math optimization libraries such as BLAS;
    \item Convolution Slicing Analysis (CSA) (\rssec{CSA}), a generic compiler analysis pass that can be used to determine a tiling strategy for convolution in different types of CPU architectures, given convolution and hardware information;
    \item Convolution Slicing Optimization (CSO) (\rssec{CSO}), a code-generation pass that results in a direct convolution \textit{macro-kernel} that improves performance when compared to Im2Col+BLAS for two CPU architectures (Intel x86 and IBM POWER10) using specialized micro-kernels;
    \item Vector-Based Packing (VBP) (\rssec{stride1-pack}), an input-tensor packing solution that leverages shift instructions in vector registers for better performance in unitary stride convolutions. 
\end{compactitem}

\rsec{overview} gives a quick overview of the compilation and execution flows of the proposed approach. \rsec{micro-kernel} describes the design of the micro-kernel using the Intel x86 and the IBM POWER10 MMA architectures. \rsec{packing} explains how packing is performed for the input image and filters. \rsec{experiments} presents a comparative experimental evaluation. 

\section{A Background on Convolution} 
\label{sec:background}

This section introduces convolution, matrix-multiplication acceleration extensions, and the notation used in this paper.

\subsection{Convolution Operation}

Convolution in CNNs is formulated as a matrix operation between a set of filters and an input tensor. The filters are also called sets of weights because they are the parameters learned during the training of convolution layers.

A convolution layer of a CNN consists of an input tensor, a set of filters, and one output tensor. In a three-dimensional convolution, the dimensions are the number of channels, height, and width of the tensors. The output tensor $\mathit{OUT}$ has dimensions $\mathit{OUT}_c \times \mathit{OUT}_h \times \mathit{OUT}_w$. 
The input tensor $\mathit{IN}$ has dimensions $\mathit{IN}_c \times \mathit{IN}_h \times \mathit{IN}_w$, and the set of $\mathit{OUT}_c$ filters ($\mathit{FS}$) has dimensions $\mathit{IN}_c \times F_h \times F_w$, where $\mathit{IN}_c$ is the number of channels, and $F_h$ and $F_w$ are the height and width of each channel in the filter. Each output channel is the result of a convolution with a different filter. 

\begin{figure*}[t]
    \includegraphics[width=0.6\linewidth]{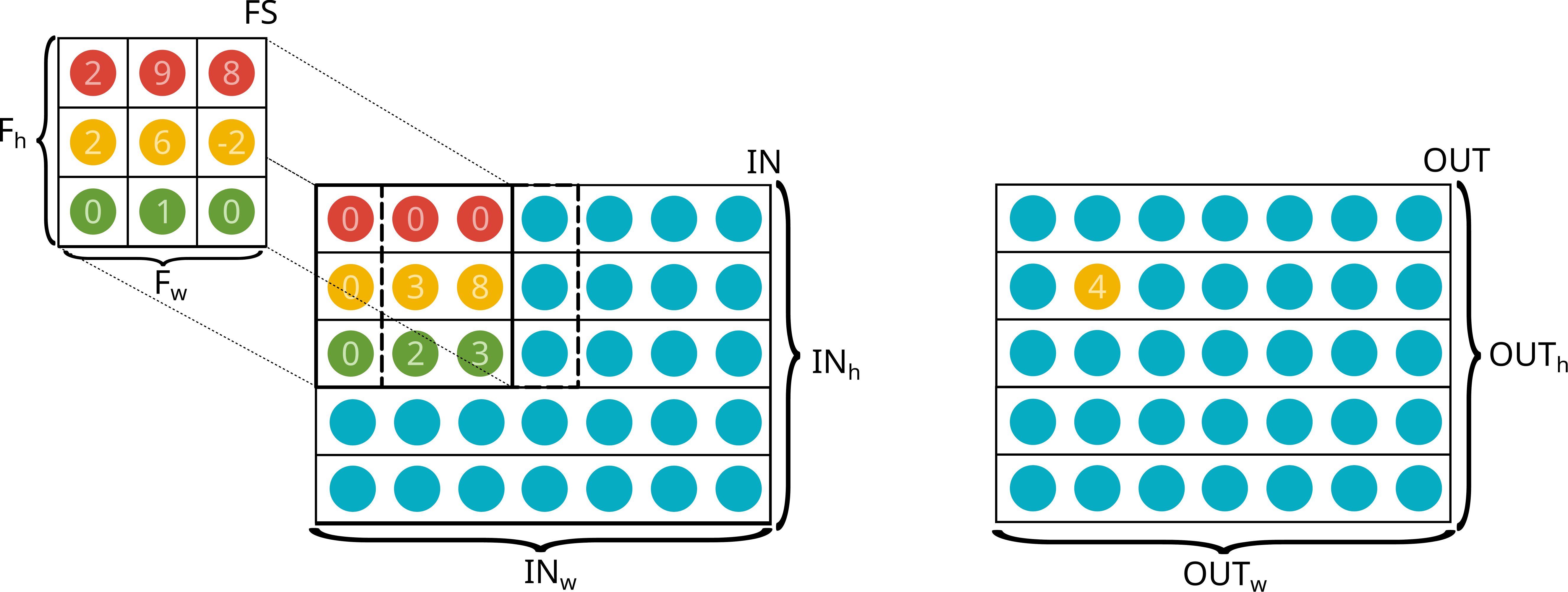}
    \caption{Convolution step with one $3 \times 3$ filter ($\mathit{OUT}_c = 1$), one channel ($\mathit{IN}_c = 1$) and unitary stride. The solid border square indicates the first window, and the dashed border square represents the second window. The convolution between the filter and the first window results in the yellow element in the output.}
    \label{fig:conv}
\end{figure*}

A window is the projection of a filter onto the input tensor, resulting in an $\mathit{IN}_c \times F_h \times F_w$ block of elements. Each output element results from the weighted sum of the input elements in a window, with the weights being the corresponding filter elements. The formation of windows from the input tensor and the computation of a convolution output element are illustrated in \rfig{conv}. Each filter slides over $\mathit{IN}$ with stride $s$ in one dimension between computations.

Tiling a convolution consists of dividing the input tensor $\mathit{IN}$, filter set $\mathit{FS}$, and output tensor $\mathit{OUT}$ into tiles that fit in the cache simultaneously.
Access to these tiles can be scheduled to take advantage of every cache level in a memory hierarchy. 

\subsection{Im2Col - Image-to-Column}

In the Im2Col + BLAS approach \cite{im2col}, an image-to-column transformation expands the input tensor into windows and then multiplies the result by the weight matrix using a highly optimized GEMM routine from a BLAS library.
The Im2Col transformation expands the input image by converting each window into a vector of elements and storing this vector as a column of a larger  $(\mathit{IN}_c \times F_h \times F_w) \times (\mathit{OUT}_h \times \mathit{OUT}_w)$ tensor. The set of filters forms an $\mathit{OUT}_c \times (\mathit{IN}_c \times F_h \times F_w)$ tensor. A GEMM operation applied to these 2D tensors produces an $\mathit{OUT}_c \times (\mathit{OUT}_h \times \mathit{OUT}_w)$ output tensor.
This approach performs two data-movement tasks:
\begin{inparaenum}[(a)]
\item the Im2Col data rearrangement; and 
\item data packing performed by the GEMM routine to increase spatial locality.
\end{inparaenum}

\subsection{POWER10 and the MMA Engine}
\label{sec:mma}

\name~uses an ISA extension for convolution operations in the IBM POWER10 architecture. This processor introduces Matrix-Multiply Assist (MMA) \cite{MMA}, a new built-in engine to accelerate tensor operations accessed through an extension of the POWER ISA v3.1. MMA uses POWER10's 128-bit vector registers (VSRs) as inputs to perform rank-$k$ outer product operations. The result is stored in one of eight new 512-bit accumulator registers representing two-dimensional matrices. The evaluation in this paper indicates that an efficient combination of MMA and VSR operations considerably improves the performance of direct-convolution operations.



\section{\name~Compilation Flow}
\label{sec:overview}

\name~is a solution to convolution that mixes compile-time data transformation for filter packing and efficient code generation to better utilize memory and architecture resources.

\begin{figure}[t]
  \includegraphics[scale=0.5, keepaspectratio]{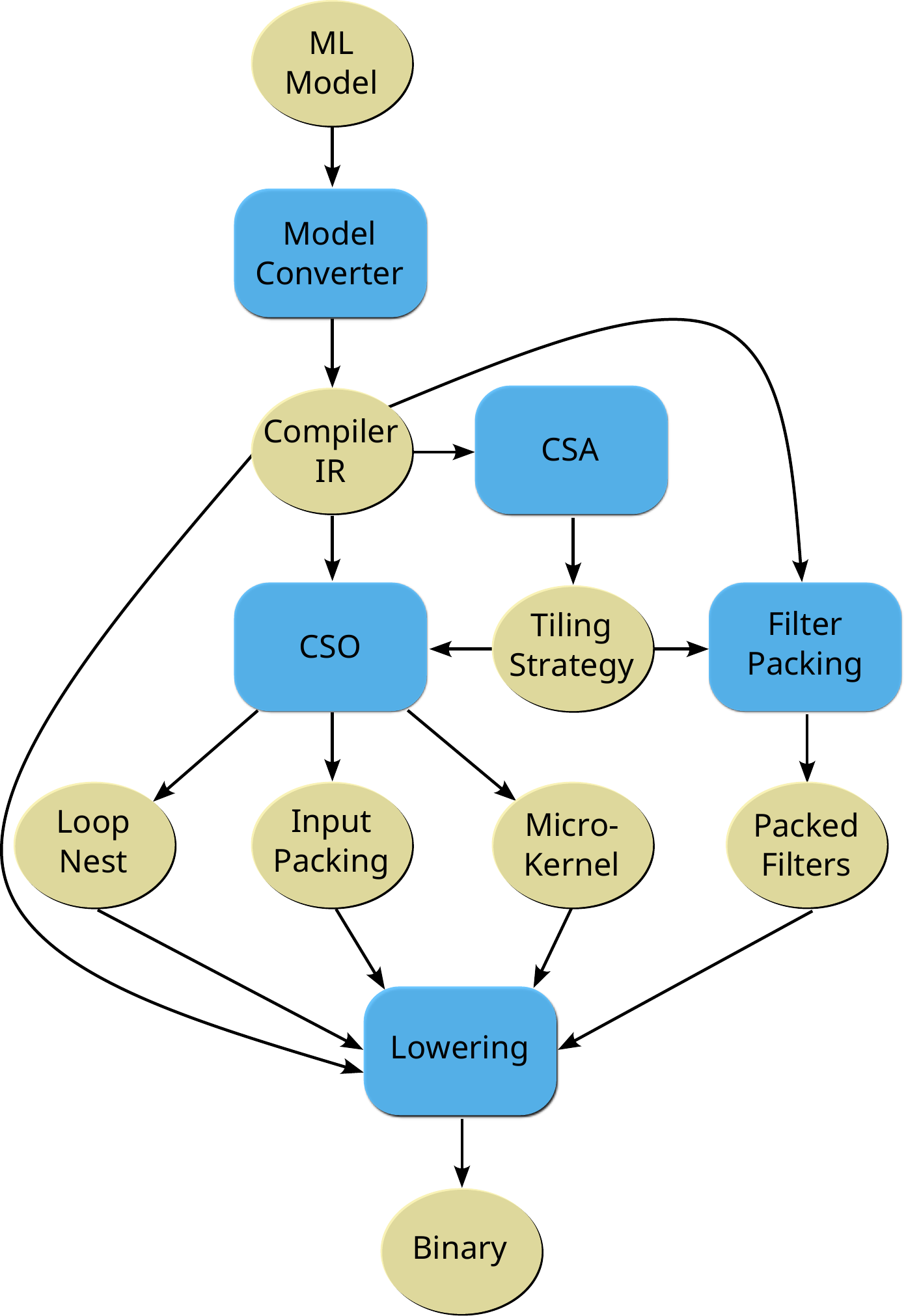}
  \caption{Compilation flow in a machine-learning compiler using \name~to optimize convolutions.}
  \label{fig:our-approach}
\end{figure}

The compilation flow in~\rfig{our-approach}, implemented using ONNX-MLIR and LLVM, can also be used in other compiling frameworks. Blocks are the computation passes and ellipses are the inputs/outputs. After the conversion of a machine-learning model into an Intermediate Representation (IR), the Convolution Slicing Analysis (CSA) algorithm determines --- for each convolution layer:
\begin{inparaenum}[(a)]
\item the sizes of the data tiles; 
\item the tile scheduling, \tit{ i.e.}, the order in which they are loaded at execution time; and
\item the number of tiles that are loaded in each step. 
\end{inparaenum}
Together, (b) and (c) determine the tile distribution in the cache hierarchy, which form a \textit{Tiling Strategy}.

Based on the Tiling Strategy, the Convolution Slicing Optimization (CSO) generates a Loop-Nest macro-kernel for the tiled convolution; an outer-product-based Micro-Kernel to compute a single output tile at peak performance for the architecture; and an Input-tensor Packing routine to rearrange each tile to the required format for the micro-kernel.

For model inference, the filter data is static and filters can be packed at compile-time based on the tiling strategy. The Lowering pass combines the information generated for each convolution with the rest of the compiler IR to produce an executable binary with tiled convolutions. For training, filter data is not static and filters are packed at execution time in the macro-kernel.

The four main components of \name~are \textbf{CSA} (\rssec{CSA}), \textbf{CSO} (\rssec{CSO}), \textbf{Input and Filter Packing} (\rsec{packing}) and the \textbf{Micro-Kernel} (\rsec{micro-kernel}). The micro-kernel and packing routines are specific to the target architecture.
CSA and CSO use architectural information to create a good tiling strategy and to generate code that implement it. The micro-kernel is designed to leverage ISA acceleration extensions and dictates the packing layout and some tiling dimensions.

\section{Convolution Slicing}
\label{sec:conv-slicing}

An efficient cache-tiling solution that aims to minimize reuse distance is at the core of a direct-convolution algorithm. \name~accomplishes efficient tiling through two compilation passes: Convolution Slicing Analysis (CSA) is responsible for computing tile sizes, distribution, and scheduling; and Convolution Slicing Optimization (CSO) generates the macro-kernel used to compute the tiled convolution based on the tiling strategy determined by CSA.

\subsection{Convolution Slicing Analysis (CSA)}
\label{sec:CSA}

Convolution Slicing Analysis (CSA), a cache-blocking-based algorithm, uses a symbolic simulation heuristic to find a suitable data tiling and scheduling combination that reduces convolution execution time. 
It aims to maximize input-tensor and filter data reuse on cache memories by: 
\begin{inparaenum}[(a)]
\item determining the tile size and distribution through the cache hierarchy that maximize cache usage for every level; and
\item selecting the order in which input-tensor/filter tiles are accessed to reduce data movement between the various levels of the cache hierarchy. 
\end{inparaenum}
In this paper $\mathit{IN}^T, \mathit{FS}^T$, and $\mathit{OUT}^T$ represent, respectively, the tiles of $\mathit{IN}$, $\mathit{FS}$, and $\mathit{OUT}$.

\begin{figure}
    \centering
    \includegraphics[width=\linewidth]{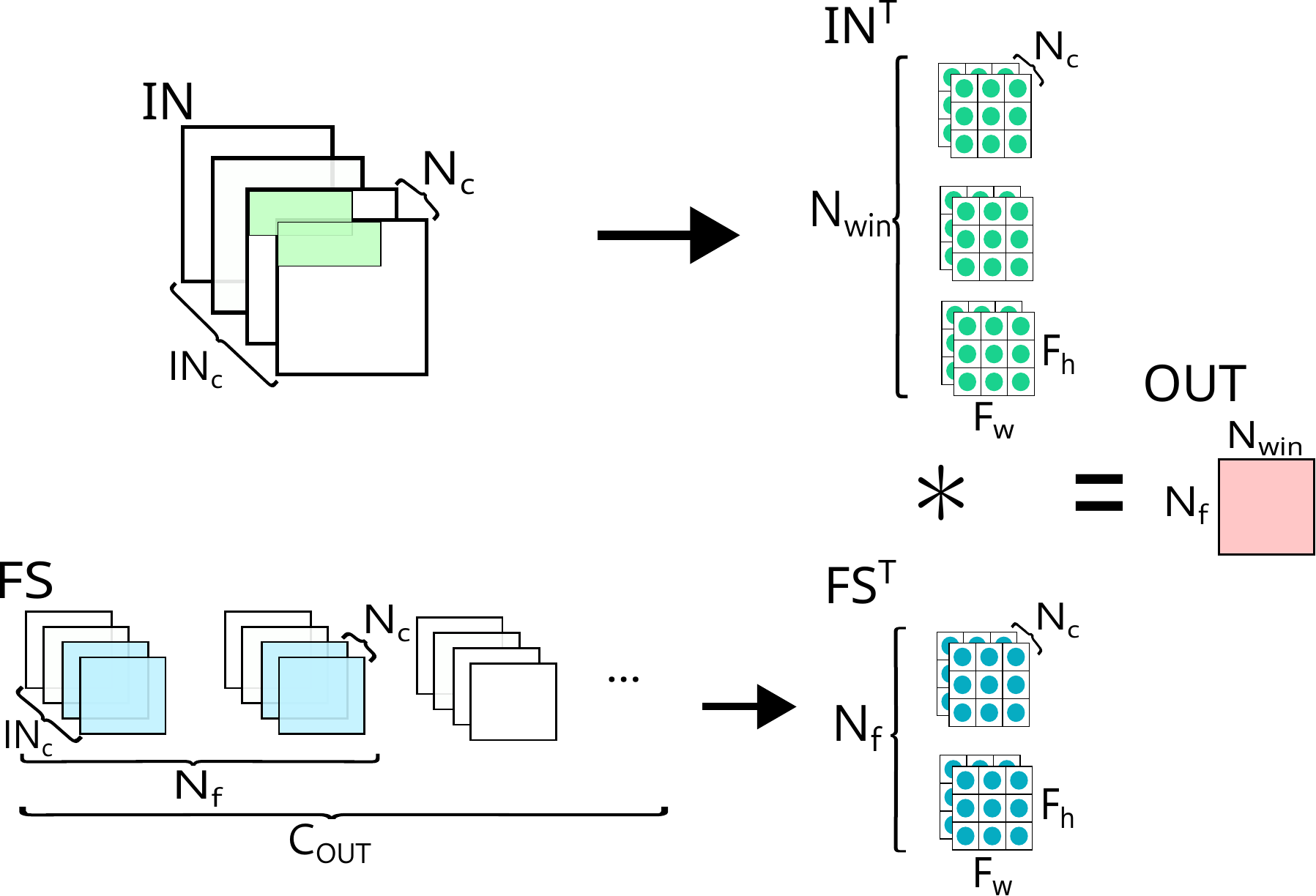}
    \caption{A CSA-tiled convolution, for $F_h = F_w = 3$, before packing.}
    \label{fig:tiles}
\end{figure}

Each convolution output element is related to an input-tensor window and there is usually an overlap between windows. Thus, each $\mathit{IN}^T$ contains multiple windows to leverage micro-kernel vectorization. CSA determines the number of windows available in each $\mathit{IN}^T$ and the number of filters in each $\mathit{FS}^T$. The tile sizes are given by \req{tiles_size}, where, for example, $\lvert\mathit{IN}^T\rvert$ represents the size of the input tile $\mathit{IN}^T$.
\rfig{tiles} shows that $N_\mathit{win}$ is the number of windows in $\mathit{IN}^T$ and $N_f$ is the number of filters used to form $\mathit{FS}^T$. 
The values of $N_\mathit{win}$ and $N_f$ are a micro-kernel design choice that aims to 
maximize performance based on the hardware that supports the computation (\rsec{micro-kernel}). 

The CSA algorithm selects tile sizes based on the following constraints:
\begin{compactitem}
    \item the windows in $\mathit{IN}^T$ and the filters in $\mathit{FS}^T$ are only tiled in their channel dimensions;
    \item full $F_h \times F_w$ filters and windows are used in a tile;
    \item the maximum number of channels ($N_c$) is chosen for each tile.
\end{compactitem}
The combination of $N_\mathit{win}$ and $N_f$ forms $\mathit{OUT}^T$, as shown in \req{tiles_size} and \rfig{tiles}.  $\mathit{DT}$ is the data type size in bytes (\tit{e.g.}, $4$ for float32).
Within the constraints above, the CSA algorithm must ensure that a $\mathit{IN}^T$, an $\mathit{FS}^T$ and an $\mathit{OUT}^T$ must fit in L1 cache. 

\begin{eqnarray}
    \lvert\mathit{IN}^T\rvert &=& N_\mathit{win} \times N_c \times F_h \times F_w  \times \mathit{DT}
   \nonumber \\
    \lvert\mathit{FS}^T\rvert &=& N_f \times N_c \times F_h \times F_w  \times \mathit{DT} \label{eq:tiles_size}
    \\
    \lvert\mathit{OUT}^T\rvert &=& N_\mathit{win} \times N_f \times \mathit{DT} \nonumber
\end{eqnarray}

$N_c$ specifies the number of input channels in a \textit{channel set}.
When $N_c = \mathit{IN}_c$ all channels are included in a single $\mathit{IN}^T$ and in a single $\mathit{FS}^T$, and thus $\mathit{OUT}^T$ can be computed in a single step.
When $N_c < \mathit{IN}_c$ each tile combination computes a partial result for an $\mathit{OUT}^T$, onto which the results from the rest of the channels need to be accumulated. In this case, the data of $\mathit{OUT}^T$ may be evicted from the L1 cache and then loaded again to continue the accumulation. 
CSA tries to avoid eviction by maximizing $N_c$ within the constraints in \req{tiles_constraint}, \tit{i.e.}, the total size of $\mathit{IN}^T$, $\mathit{FS}^T$, and $\mathit{OUT}^T$ has to be less than the available space in the L1 cache --- allowing for the space needed for prefetching and intermediate computation, \tit{e.g.} packing.  In \req{tiles_constraint}, the fraction of the L1 cache available for tile storage is represented by $\alpha$, which depends on the hardware and can be tuned to each convolution. An alternative formulation could  subtract the estimated space for prefetching and intermediary values from $|\mathit{L1}|$.

\begin{align}
\label{eq:tiles_constraint}
\begin{split}
    \lvert\mathit{IN}^T\rvert + \lvert\mathit{FS}^T\rvert + \lvert\mathit{OUT}^T\rvert \leq \lvert\mathit{L1}\rvert \times \alpha
\end{split}
\end{align}

CSA calculates the number of $\mathit{IN}^T$, $\mathit{FS}^T$, and $\mathit{OUT}^T$ required to compute the convolution. This calculation is done individually for each channel set.
\begin{eqnarray}
\label{eq:number_of_tiles}
    \#\mathit{IN}^T &=& \frac{\mathit{OUT}_h \times \mathit{OUT}_w}{N_\mathit{win}}  \nonumber \\
    \#\mathit{FS}^T &=& \frac{\mathit{OUT}_c }{ N_f} \\
    \#\mathit{OUT}^T &=& \#\mathit{IN}^T \times \#\mathit{FS}^T   \nonumber
\end{eqnarray}
\textit{Edge cases} arise when the tiling values are not proper divisors of the convolution information. For instance, when $\mathit{IN}_c$ is not divisible by $N_c$. These cases are handled by computing smaller tiles when necessary, with smaller $N_c$, $N_\mathit{win}$ and/or $N_f$ values.

CSA relies on two possible execution orders when scheduling tiles:
\begin{inparaenum}[(a)]
\item \textit{Input Stationary (IS)}, keeps one $\mathit{IN}^T$ stationary in L1 cache and reuses it over as many $\mathit{FS}^T$ as possible before proceeding to the next $\mathit{IN}^T$; 
\item \textit{Weight Stationary} (WS), keeps one $\mathit{FS}^T$ stationary in L1, using it to compute multiple $\mathit{OUT}^T$ from many $\mathit{IN}^T$, before moving to the next $\mathit{FS}^T$.
\end{inparaenum}

\begin{figure*}[t]
  \includegraphics[width=0.7\linewidth]{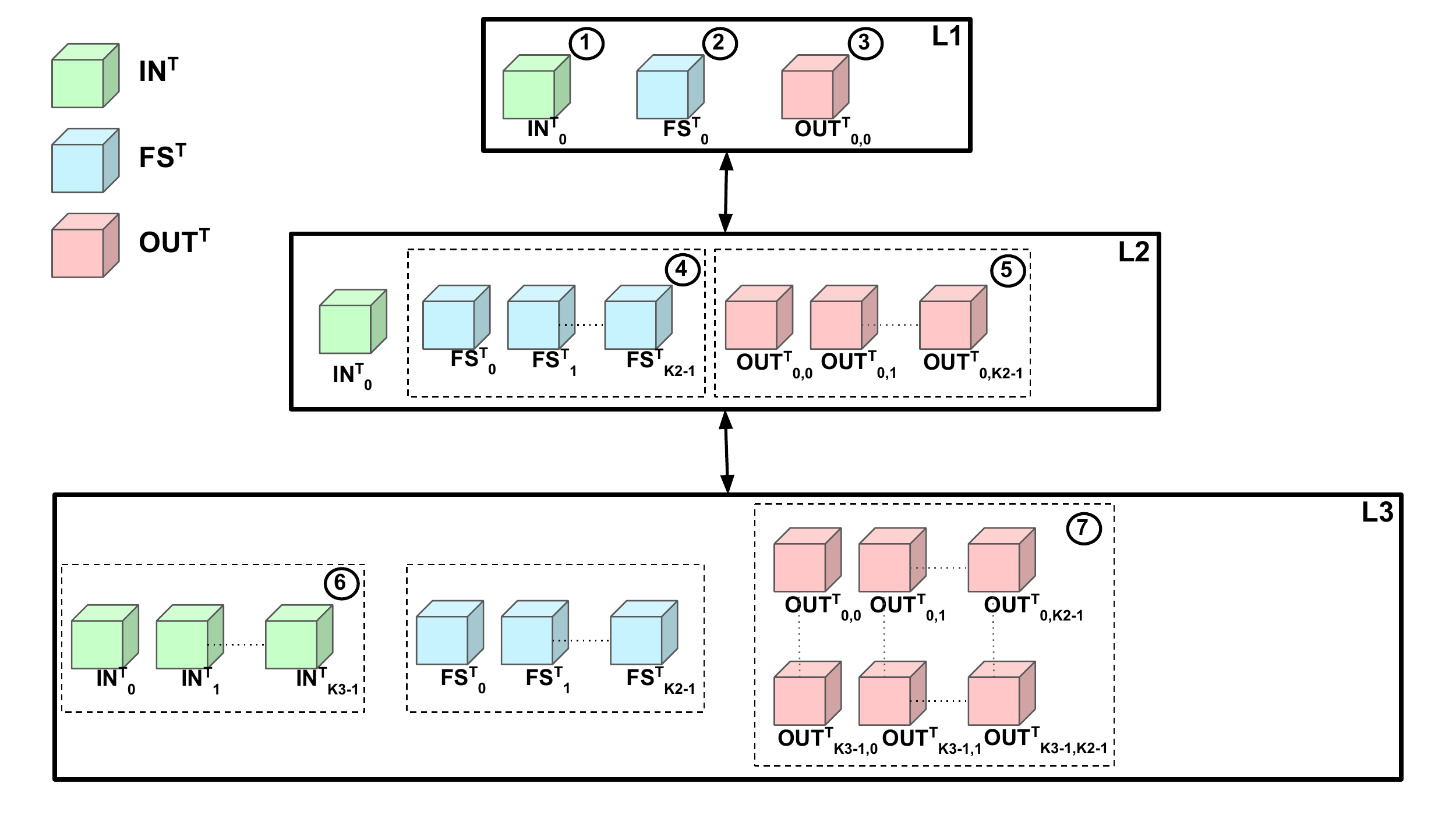}
  \caption{CSA tile distribution over a three-level cache hierarchy with Input Stationary scheduling.} 
  \label{fig:csa_is}
\end{figure*}

Both the IS and the WS scheduling strategies lead to the specification of $\mathit{K2}$, the number of tiles in the L2 cache, and of $\mathit{K3}$, the number of tiles in the L3 cache. These values are set to use the storage available in each cache level while increasing reuse. The L1 cache must always contain one set of  $\mathit{IN}^T$, $\mathit{FS}^T$, and $\mathit{OUT}^T$, which are used for the micro-kernel computation. 

\rfig{csa_is} illustrates the data movements and computation for the IS schedule. Initially, an $\mathit{IN}^T$ is loaded to L1 \circled{1} and kept stationary. Next, a $\mathit{FS}^T$ is loaded to L1 \circled{2} so that the micro-kernel is executed to compute an $\mathit{OUT}^T$ \circled{3}, which must also fit into L1. While the $\mathit{IN}^T$ remains in L1, $(\mathit{K2} - 1)$  $\mathit{FS}^T$ \circled{4} are in turn brought to L1 to compute other $(\mathit{K2} - 1)$ $\mathit{OUT}^T$ \circled{5}. The maximum value of $\mathit{K2}$ that satisfies the inequality in \req{is_k2} ensures that the stationary $\mathit{IN}^T$ and the $\mathit{K2}$ $\mathit{FS}^T$ and $\mathit{OUT}^T$ fit into the L2 cache. The $\mathit{K2}$ $\mathit{FS}^T$ kept in L2 are reused by $(\mathit{K3} - 1)$ other $\mathit{IN}^T$ that are kept in L3 \circled{6}. The maximum value of $\mathit{K3}$ that satisfies the restriction in \req{is_k3} ensures that a set of $\mathit{K3}$ $\mathit{IN}^T$, $\mathit{K2}$ $\mathit{FS}^T$ and the results they generate \circled{7} fit into the L3 cache. 

The number of times that each tile is loaded to L1 is given as follows:
\begin{compactitem}
    \item If $\mathit{K2} = \#\mathit{FS}^T$, then loading each $\mathit{IN}^T$ to L1 one time is enough. Otherwise, other $\mathit{K2}$ $\mathit{FS}^T$ are loaded from memory to L2, and $\mathit{K3}$ $\mathit{IN}^T$ from L3 are loaded to L1 again; 
    \item If $\mathit{K3} \neq \#\mathit{IN}^T$, then the same $\mathit{K2}$ $\mathit{FS}^T$ are loaded back to L2. In this case, other $\mathit{K3}$ $\mathit{IN}^T$ are loaded to L3 and these steps repeat; 
    \item The $\mathit{OUT}^T$ are loaded back to L1 for accumulation $((\mathit{IN}_c / N_c) - 1)$ times. 
\end{compactitem}

In WS scheduling, the order of $\mathit{IN}^T$ and $\mathit{FS}^T$ are reversed. The share of the caches available to be used in the equations that define $\mathit{K2}$ and $\mathit{K3}$ are represented in \reqs{is_k2}{is_k3} by $\beta$ and $\gamma$. Alternatively, the estimated space needed for packing, prefetching, and other operations, can be subtracted from the cache size.

\begin{eqnarray}
    \lvert\mathit{IN}^T\rvert + \mathit{K2} \times (\lvert\mathit{FS}^T\rvert + \lvert\mathit{OUT}^T\rvert) \leq \lvert\mathit{L2}\rvert \times \beta \label{eq:is_k2}\\
    \mathit{K3} \times \lvert\mathit{IN}^T\rvert + \mathit{K2} \times \lvert\mathit{FS}^T\rvert + \mathit{K2} \times \mathit{K3} \times \lvert\mathit{OUT}^T\rvert \leq \lvert\mathit{L3}\rvert \times \gamma \label{eq:is_k3}
\end{eqnarray}

Exploring the tiling space to find the optimal strategy might lead to very long compilation times. Contrary to other approaches, such as TVM \cite{tvm} or Ansor \cite{ansor}, that consider long compilation and code generation times acceptable (hours), we believe that ML model development  cycle should be fast enough (dozens of minutes) to enable  the designer to explore many different model architectures. Hence, CSA uses a simple tiling-space exploration heuristic, applied to $N_c$, $\mathit{K2}$, and $\mathit{K3}$, that has been shown to produce quality code at reasonable compilation times. Each parameter starts with its highest possible value \tit({e.g.} $\mathit{IN}_c$ for $N_c$), and is halved at each following iteration until all constraints are satisfied. Exploring other tiling-space exploration strategies is left for future work.

CSA decides between IS and WS based on a cost model (\rsec{costmodel}) using the number of L2, L3, and main memory accesses and estimating the number of cycles for those accesses in each strategy.

\subsubsection{Cost Model}
\label{sec:costmodel}

$C_\mathit{TOTAL}$ is the cost of loading tiles from L2 cache, L3 cache, and main memory.
It is the sum of the product of the number of loads from each level and the corresponding cost (in cycles) of each load/store operation at that level (\req{cost}).   
The cost of an L1 hit is not included because every load operation goes through the L1 cache. 
The cost of reloading an output tile when $N_c \neq \mathit{IN}_c$ is also not included in the model because it is the same for both the IS and the WS strategies. 

\begin{eqnarray}
    C_\mathit{TOTAL} = C_\mathit{DRAM} \times N_\mathit{DRAM} + C_\mathit{L3} \times N_\mathit{L3} + C_\mathit{L2} \times N_\mathit{L2}\label{eq:cost}\\
    N_\mathit{DRAM} = N_{\mathit{DRAM}_1} + N_{\mathit{DRAM}_2}\label{eq:dram-cost}
\end{eqnarray}

The number of DRAM accesses (\req{dram-cost}) is the sum of $N_{\mathit{DRAM}_1}$, the number of \textit{cold misses} required to touch every single tile at some point of the convolution execution; and $N_{\mathit{DRAM}_2}$, additional touches needed to reload tiles if they are not available in the cache when required. 


Let $\mathit{CL}$ be the cache line size of the current architecture. The number of cold misses is given by:

\begin{align}
    N_{\mathit{DRAM}_1} = \frac{\mathit{IN}_c}{N_c} \times \left(\frac{\#\mathit{IN}^T \times \lvert\mathit{IN}^T\rvert + \#\mathit{FS}^T \times \lvert\mathit{FS}^T\rvert}{\mathit{CL}}\right)\label{eq:dram1-cost}
\end{align}

All other equations in this section depend on the analyzed scheduling strategy. The rest of this section will consider the Input Stationary case. For the Weight Stationary cost equations, replace $\#\mathit{IN}^T$ and $\lvert\mathit{IN}^T\rvert$ with $\#\mathit{FS}^T$ and $\lvert\mathit{FS}^T\rvert$ and vice-versa.

When all the $\mathit{FS}^T$ do not fit simultaneously in the L2 cache, the reloading of tiles during the computation for each set of $\mathit{K3}$ $\mathit{IN}^T$ leads to additional DRAM accesses. This data is not available in the L3 cache because it is evicted when loading the other $\mathit{IN}^T$ and $\mathit{FS}^T$ sets, that use all the available space.
\reqs{fs-fit}{dram2-cost} computes $N_{\mathit{DRAM}_2}$,  the number of additional accesses required. 
 
\begin{gather}
    \mathit{FS}^T_{\mathit{fitmin}} = \min\left(\left(\frac{\#\mathit{FS}^T}{\mathit{K2}} - 1\right), 1\right)\label{eq:fs-fit}\\
    \mathit{IN}^T_{\mathit{fit}} = \frac{\#\mathit{IN}^T}{\mathit{K3}} - 1\label{eq:in-fit}\\
    N_{\mathit{DRAM}_2} = \frac{\mathit{IN}_c}{N_c} \times \frac{\mathit{FS}^T_{\mathit{fitmin}} \times \mathit{IN}^T_{\mathit{fit}} \times \#\mathit{FS}^T \times \lvert\mathit{FS}^T\rvert}{\mathit{CL}}\label{eq:dram2-cost}
\end{gather}              
     
$\mathit{FS}^T_\mathit{fitmin}$  represents the condition for additional DRAM accesses, \tit{i.e.} if $\#\mathit{FS}^T$ fits in the L2 cache. If the equation evaluates to zero, no reloads are needed. Otherwise, the number of loads from memory is controlled by $\mathit{IN}^T_\mathit{fit}$, the number of additional $\mathit{IN}^T$ sets.
     
An $\mathit{IN}^T$ may need to be loaded from the L3 cache more than once. If there is more than one set of $\mathit{K2}$ $\mathit{FS}^T$, the next set will need to be computed with all $\mathit{K3}$ $\mathit{IN}^T$ in the L3 cache, so they have to be loaded to L1 again.
In the calculation of the number of loads into L3 (\reqs{fs-fit2}{l3-cost}), $\mathit{FS}^T_\mathit{fit}$ controls the number of reloads from the L3 cache with the number of additional $\mathit{FS}^T$ sets. 

\begin{eqnarray}
    \mathit{FS}^T_\mathit{fit} &=& \frac{\#\mathit{FS}^T}{\mathit{K2}} - 1\label{eq:fs-fit2}\\
    N_\mathit{L3} &=& \frac{\mathit{IN}_c}{N_c} \times \frac{\mathit{FS}^T_\mathit{fit} \times \#\mathit{IN}^T \times \lvert\mathit{IN}^T\rvert}{\mathit{CL}}\label{eq:l3-cost}
\end{eqnarray}

For every $\mathit{IN}^T$, all $\mathit{K2}$ $\mathit{FS}^T$ need to be loaded from the L2 cache for computation. Since the tiles initially come from memory directly to the L1 cache, one access should be discharged. $N_\mathit{L2}$ is calculated using \req{l2-cost}.

\begin{align}
    N_\mathit{L2} = \frac{\mathit{IN}_c}{N_c} \times \frac{\left(\#\mathit{IN}^T - 1\right) \times \#\mathit{FS}^T \times \lvert\mathit{FS}^T\rvert}{\mathit{CL}}\label{eq:l2-cost}
\end{align}

The cost model represents all data movement between cache levels and main memory, and they depend on the scheduling strategy. A lower cost means better tile reuse and cache usage. Thus, the cost should be calculated for both approaches, and the lowest value should be chosen, for each convolution, for that specific architecture.

\subsection{Convolution Slicing Optimization (CSO)}
\label{sec:CSO}

\begin{figure}[ht]
    \centering
    \includegraphics[height=0.68\textheight]{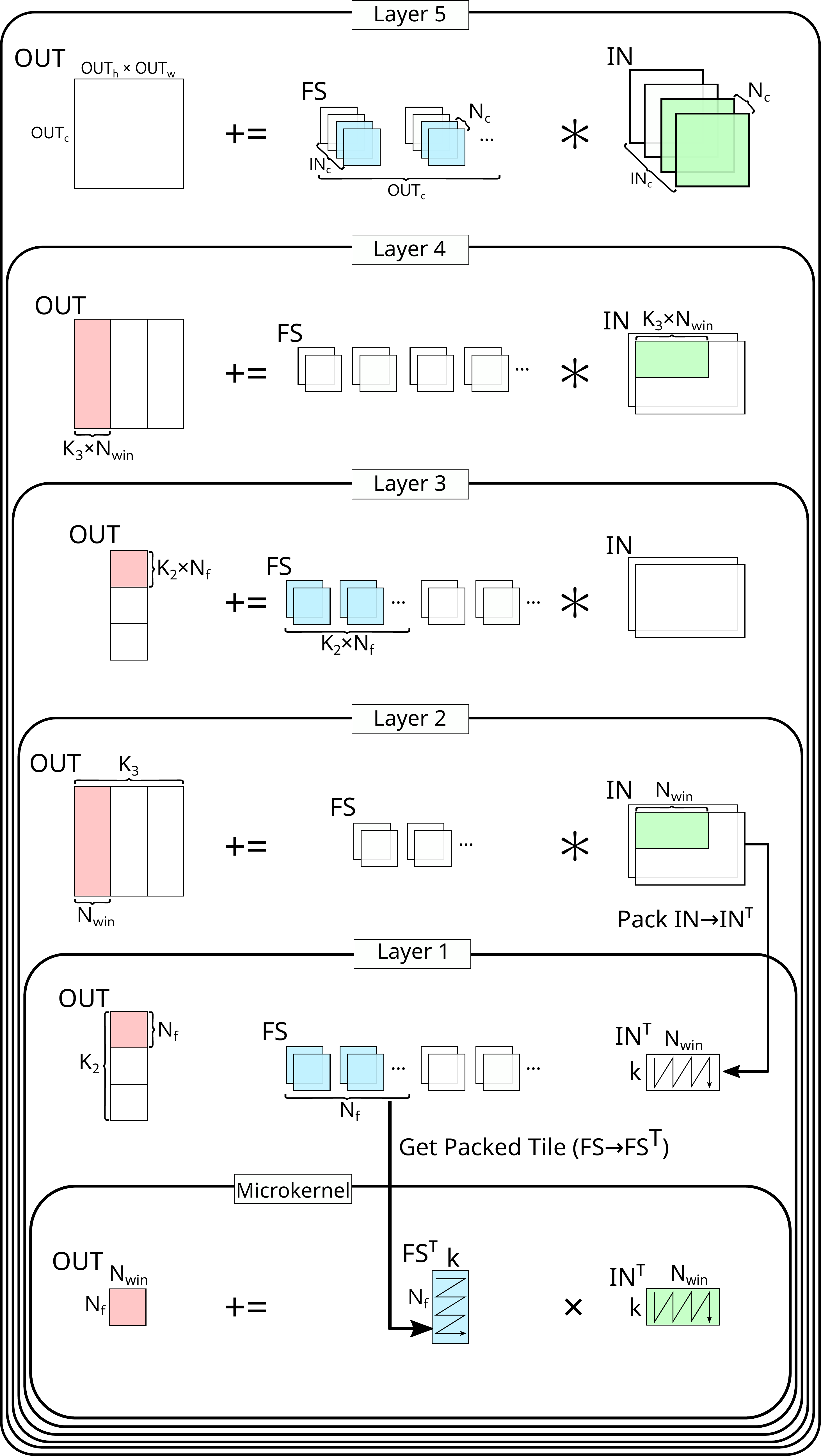}
    \caption{Convolution macro-kernel generated by CSO using CSA's tiling parameters. Illustrates the process for Input Stationary scheduling.}
    \label{fig:macro-kernel}
\end{figure}


The Convolution Slicing Optimization (CSO) slices the input, filters, and output based on the tiling sizes and scheduling computed by CSA and generates a macro-kernel to compute the tiled convolution. The main idea is to maximize tile reuse to take advantage of the cache hierarchy. 

The input-tensor packing routine (\rssec{input_pack}) expands the input into windows for one tile and packs the tile such that the elements are in the order required by the micro-kernel.  The CSO macro-kernel calls the input-tensor packing routine and iterates over all tiles in the order established by CSA. The loop nest is represented as layers for Input Stationary scheduling in \rfig{macro-kernel}. This section details that figure.

For efficiency, the micro-kernel in the innermost loop (\rsec{micro-kernel}) computes with tiles already loaded into the L1 or L2 caches.  Thus, the algorithm implements \tit{Packing on Demand}: layer two of the loop nest packs input-tensor tiles right before they are used. 
The filter-packing routine, used in layer 1, returns the already packed tile since the filters have already been packed during the model's compilation.


The outermost loop (layer five) iterates over channel sets with no reuse between them. Once a channel set is selected, the tiles within this channel set need to be further divided so that a few stay at each cache level. This  is done via the $\mathit{K2}$ and $\mathit{K3}$ parameters. 

Layers 4 and 2 are responsible for managing all $\mathit{IN}^T$. Layer 4 iterates over different sets of $\mathit{K3}$ input-tensor tiles, and layer 2 iterates over a single set, packing one tile at a time. Similarly, Layers 3 and 1 iterate over all $\mathit{FS}^T$. Layer 1 calls the micro-kernel, so this setup ensures that every combination of $\mathit{IN}^T$ and $\mathit{FS}^T$ is computed.

The $\mathit{IN}^T$ selected by layer 2 is kept stationary in the L1 cache since it is used with $\mathit{K2}$ $\mathit{FS}^T$ in layer 1. When the next $\mathit{IN}^T$ is selected, these filter-set tiles are in the L2 cache to be reused. Likewise, $\mathit{IN}^T$ are kept in the L3 cache afterward for reuse with the next set of $\mathit{FS}^T$.

In Weight Stationary scheduling, the process is the same, switching $\mathit{IN}^T$ and $\mathit{FS}^T$. Since the $\mathit{IN}^T$ need to be packed for the L2 cache in this case, the packing process happens in layer 3 for all $\mathit{K2}$ tiles at once, and each tile is fetched when needed.

\section{An Outer-Product Micro-Kernel}
\label{sec:micro-kernel}

The micro-kernel, located in the most internal layer of \rfig{macro-kernel}, computes the arithmetic operations that form the convolution. For most efficiency, the micro-kernel should be treated as an extension of the hardware instead of a high-level routine. A micro-kernel implementation is tailored to the architecture and uses any available specific instructions to increase throughput. This work focuses on a single-precision (32-bit) floating-point micro-kernel.

\name~uses an outer-product-based micro-kernel. The outer product is generally useful due to its versatility for computing higher-rank operations and high throughput, computing $n^2$ output elements from $2n$ input elements. For those reasons, it is widely used in high-performance linear algebra libraries, and it is being incorporated in CPU ISA extensions such as IBM POWER10 MMA \cite{MMA}. The micro-kernel computes a sum of outer products between vectors of floating-point elements, loaded from memory with unitary stride. A highly optimized implementation of such micro-kernel is found in BLAS GEMM, which can be used as the inner layer of the algorithm, repurposed by the packing layout and macro-kernel. This brings an additional benefit: those architectures with a BLAS implementation can easily integrate their micro-kernel into \name. Moreover, a new micro-kernel designed for some future architecture could seamlessly be used by both a BLAS library and \name.


\subsection{Micro-Kernel for IBM's MMA}

For the POWER10 architecture, the MMA engine can compute multiple output elements at once. Each input element is used multiple times to compute different output elements. Thus, the input should be split into windows when packing.

\begin{figure*}
    \centering
     \subfloat[][Window and filter data distribution over a sequence of outer products.]{
        \includegraphics[width=0.7\linewidth]
        {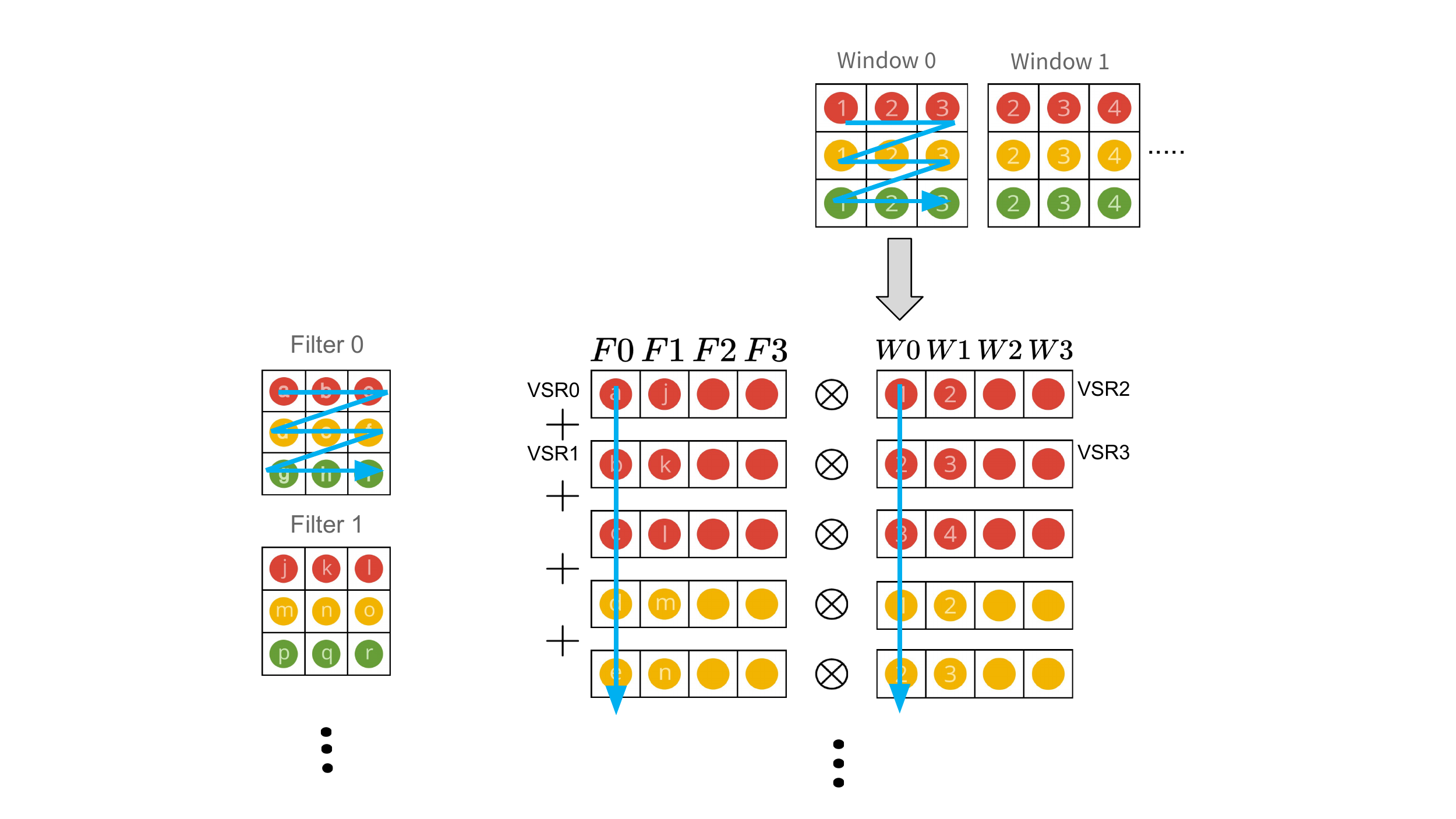}
        \label{fig:outer-product-load}
    }
    \hfill
     \subfloat[][Sequence of outer products to generate one accumulator matrix with part of the micro-kernel's convolution output.]{
        \includegraphics[width=0.6\linewidth]
        {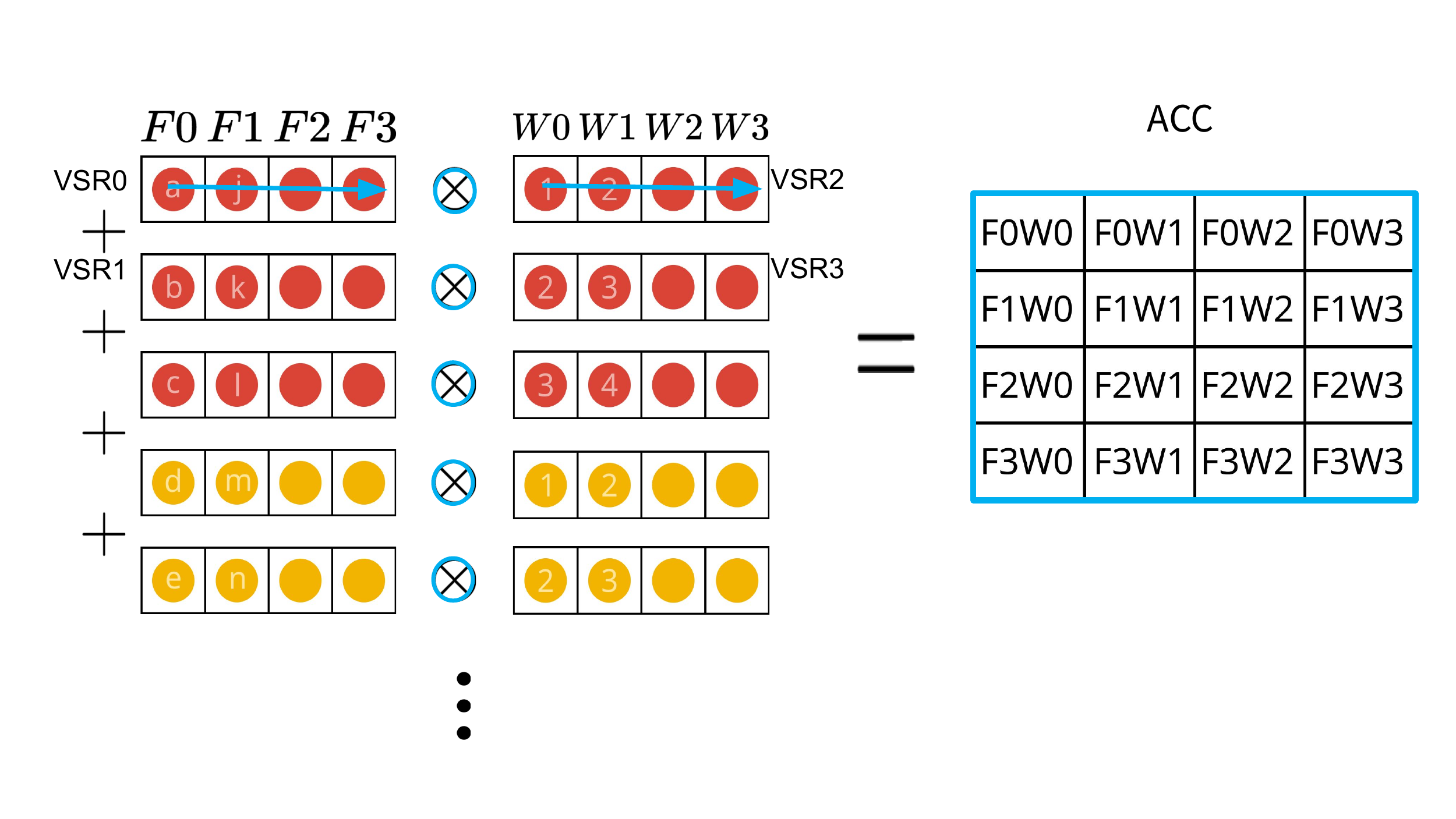}
        \label{fig:outer-product-out}
     }
    \caption{Execution of a POWER10 convolution micro-kernel for 32-bit floating point. Four filters and four windows are used in each VSR of the outer product.}
    \label{fig:outer-product}
\end{figure*}

The outer-product-based design of MMA enables vectorization of the convolution, as shown in \rfig{outer-product-load}. For 32-bit data types, each VSR contains four elements. One VSR contains elements from four different filters and the other from four different windows. An outer product between two VSRs computes sixteen partial output elements, as shown in \rfig{outer-product-out}. In the two-dimensional matrix produced, each row represents a convolution with one filter, and each column represents convolutions with one window. The partial products are accumulated in-place in this accumulator. The accumulated values are stored to memory after the complete set of $F_h \cdot F_w \cdot N_c$ outer product operations have been executed. The resulting data layout is correct for the convolution output, and thus there is no need for unpacking or reordering.

Combining all available VSRs, a larger accumulator can be emulated with a layout-dependent size. Its bigger dimension should be used for computing more windows in each micro-kernel call because there are usually more windows than filters in a convolution. This "super accumulator" has $8 \times 16$ elements in total \cite{compiling-mma} and the micro-kernel consumes eight filters and sixteen windows per call.

For POWER10 CSA uses $N_f = 8$ and $N_\mathit{win} = 16$ as described in \rssec{CSA}.

\subsection{Micro-Kernel for Intel's AVX-512}

Even though the Intel Skylake x86 architecture does not feature a matrix engine, the data layout for the input and output structures is the same as for the POWER10 micro-kernel. In this architecture, vector instructions in AVX-512 vector registers emulate the outer-product computation using the following algorithm:
\begin{compactitem}
\item load two AVX-512 registers with 4 elements (128 bits) each;
\item broadcast the elements to fill the rest of each register with copies;
\item permute the results so that copies of the same element are consecutive;
\item perform a Fused Multiply-Add (FMA) operation between the registers. 
\end{compactitem}
This micro-kernel computes $16$ windows ($N_\mathit{win}$) and $24$ filters ($N_f$) per call. A similar process can be done to emulate outer products in other architectures with SIMD computing.

\section{Packing on Demand for the Micro-Kernel}
\label{sec:packing}

Since the micro-kernel is responsible for computing the convolution and is the innermost step in the loop nest, it should be the focus for maximizing performance. This entails minimizing CPU stalls and data-manipulation overhead. Therefore, the required data should be stored sequentially in memory, in the correct order, and ideally available in cache. Therefore, \name~includes packing steps for both the input tensor and filters, as indicated in layers 2 and 1 from \rfig{macro-kernel}, respectively.

The objective of the packing steps is to reshape and, in the input-tensor case, to expand a tile of data into the format required by the micro-kernel. Thus, this format can differ between architectures. The packing routines are called for a single tile at a time and the data is loaded into the L1 cache for the micro-kernel to use thereafter (Packing on Demand).

\subsection{Input-Tensor Packing}
\label{sec:input_pack}

The input must be expanded into windows, which are then reshaped so that the micro-kernel can be used to compute the convolution. Instead of doing this process in two separate steps as done by Im2Col+BLAS,  the \name~input-tensor packing routine combines both into a single pass.

\begin{figure*}
    \centering
    \includegraphics[width=0.6\linewidth]{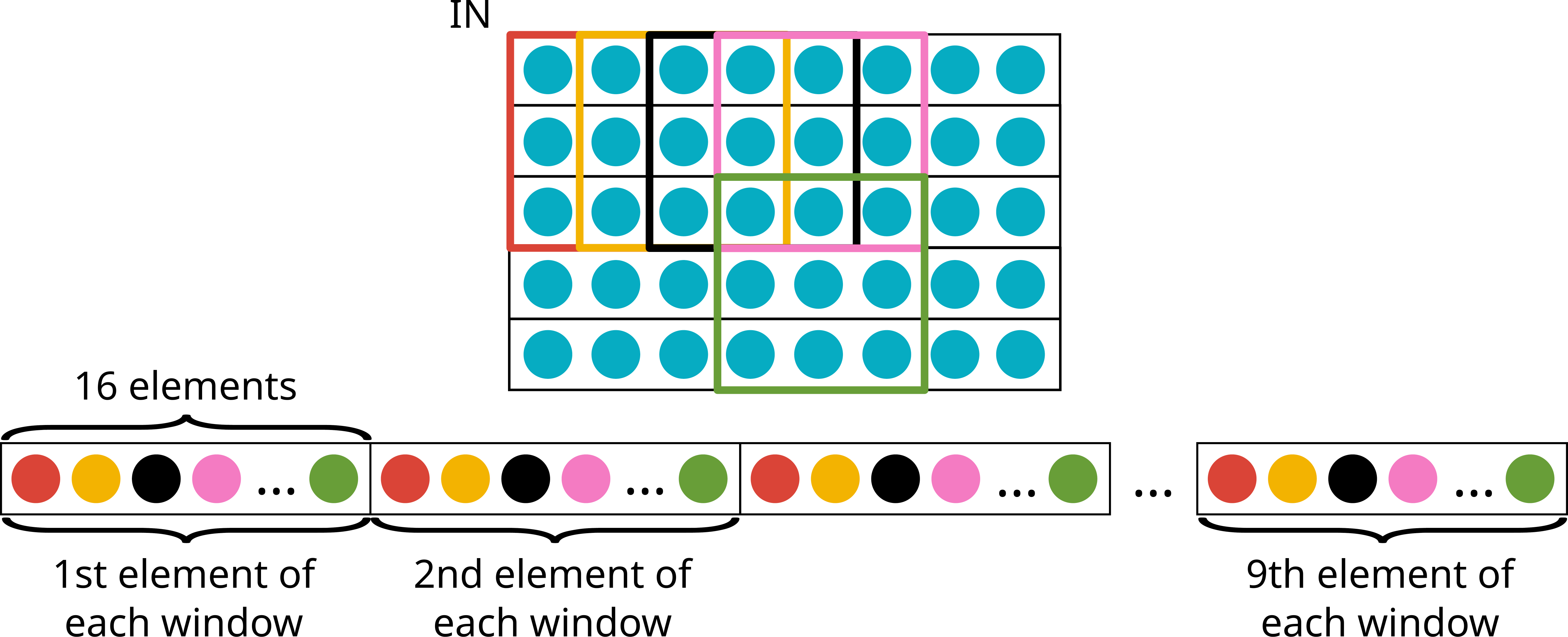}
    \caption{\name's input-tensor packing layout for a 16-window micro-kernel, applied on a convolution with $F_h=F_w=3$, $\mathit{IN}_c=1$ and unitary stride. For each of the 16 windows, the elements from the same position are stored sequentially.}
    \label{fig:packing}
\end{figure*}

\rfig{outer-product-load} illustrates the micro-kernel's access pattern in the POWER10 architecture. A window's elements are distributed between multiple outer products, so a vector register contains elements from the same position in different windows. In \rfig{outer-product}, VSR2 contains the first element of each window, VSR3 has the second element of each window, and so on. Therefore, the elements in memory should follow this pattern. Other outer-product-based micro-kernels, such as the one for Intel Skylake x86, follow this same access pattern.

\rfig{packing} shows the packing layout derived from the micro-kernel's access pattern. Elements are loaded multiple times. Thus, the data is packed so that the elements from the same position in every window are stored sequentially in memory. This pattern is repeated for all $N_c$ channels in the tile so that the packing routine takes advantage of both spatial and temporal data locality per row, especially in unitary stride convolutions.  

\subsection{Vector-Based Input Packing}
\label{sec:stride1-pack}

Packing the input tensor in \name~requires data replication because one element can be a part of many windows. In fact, most elements in an input tensor participate in many windows. 

The most straightforward way to optimize for multiple loads of the same element is to load from the L1 cache whenever possible. However, when the convolution has a unitary stride, the first element from the next window is the second element from the current window, excluding border cases (as evidenced by \rfig{packing}).
Therefore, a shift-left operation with a serial-in element is enough to get the next element from each of the windows in a packed tile.


This observation enables the use of Vector-Based Packing (VBP), where a vector shift operation reduces loads: the next element of each window in a row is obtained by shifting the current set of elements to the left, which can be done by loading all elements into vector registers. This operation is available in most architectures, so the technique is general. Both POWER10 and Intel Skylake x86 provide instructions to shift a pair of vector registers together to the left with the second vector register appended to the right of the first. The least significant  element of the first vector register is fed by the  most significant element of the second vector register.

Padding the input tensor at the start of the convolution process addresses edge cases. Convolutions with $1 \times 1$ filters can also benefit from vector register use, even if no shifts are needed.

\subsection{Filter Packing} 
\label{sec:filter-pack}

Filters are loaded in the same way as windows in the micro-kernel. Thus, the packing strategy is similar: a position from each tile filter is packed before advancing to the next, and this process repeats for every channel. However, in contrast with the input tensor, there is no data replication while packing the filters, only the rearrangement of their elements.

Filter packing differs between the two tested architectures because their micro-kernels compute a different number of filters per call. 

\section{Comparing \name~With Im2Col + BLAS}
\label{sec:experiments}

This experimental evaluation indicates that the strengths of \name~come from the use of faster micro-kernels that are tailored for each architecture, and from the careful packing that reduce cache misses.  


\subsection{Full-Model Performance Evaluation Using ONNX-MLIR}
\label{sec:setup}

The \name~prototype evaluated in the two CPU architectures specified in~\rtab{archs} is integrated into the ONNX-MLIR framework \cite{onnx-mlir} to allow for full-model performance evaluation under realistic conditions. 
The loop nest, input-tensor packing routine and micro-kernel are implemented as a library instead of being completely generated by CSO at compile-time. 
All the $393$ convolutions found in seven machine-learning models from the ONNX Model Zoo \cite{modelzoo}, are executed with \name. 
Each model takes one input image, thus the batch is 1. 

\begin{table}[!t]
\centering
\resizebox{\columnwidth}{!}{%
\begin{tabular}{|c|c|c|}
\hline
\multicolumn{1}{|c|}{\multirow{1}{*}{\textbf{Name}}} & \multicolumn{1}{c|}{\textbf{IBM Power E1050}} & \multicolumn{1}{c|}{\textbf{Intel Xeon Silver 4208}} \\ \hline
{CPU Architecture} & IBM POWER10 & Intel Cascade Lake (x86) \\ 
{Matrix/SIMD Engines} & MMA/VSX & None/AVX-512 \\ 
{L1 Cache Size} & 32 kB & 32 kB \\ 
{L2 Cache Size} & 1 MB & 1 MB\\ 
{L3 Cache Size} & 4 MB & 4 MB \\ 
{CPU Frequency} & 4 GHz & 2.1 GHz \\ 
{Matrix/SIMD Units} & 2/4 & 0/1 \\ 
{Max Throughput} & 256 GFLOPS/s & 67.2 GFLOPS/s \\ \hline
\end{tabular}}
\caption{Details of CPU architectures on which tests were conducted.}
\label{tab:archs}
\end{table}

The baseline, called \base, is an industry-standard convolution algorithm where the input is packed into a patch-matrix using the Im2Col routine from the Caffe framework~\cite{caffe}, followed by a call to an optimized GEMM routine from the OpenBLAS library~\cite{openblas}. 
For a fair comparison, the micro-kernels used in \name~are from this library. 

\req{tiles_constraint} and \reqs{is_k2}{is_k3} introduce the $\alpha$, $\beta$, and $\gamma$ parameters to account for extra space needed for non-tile data in each level of the memory hierarchy. 
These parameters can be tuned for each use case. 
For this performance evaluation, the values of all three parameters were set to $0.9$ for all convolutions and both machines. 
Exploration with slight variations around these values revealed that there was no significant performance effects.

Both \name~and \base~ use single-precision floating-point datatype, are executed in single-threaded mode, and are compiled with Clang 14 for the x86 architecture and IBM XL C/C++ for Linux 17.1.1 for POWER10.
Cache performance measurements for x86 use the Perf Events API from the Linux kernel. 
All results are averaged over 100 runs, with less than 5\% observed standard deviation for both architectures.

\subsection{WS $\times$ IS Scheduling}
\label{sec:scheduling-results}

The convolution scheduling, determined by the cost model (\rssec{costmodel}), is sensitive to the micro-kernel shape ($N_f \times N_{\mathit{win}}$). The POWER10 micro-kernel computes $8 \times 16$ elements, while the x86 micro-kernel computes $24 \times 16$ elements. The change in the $N_f$ value leads to a larger $\mathit{FS}^T$, which may favor its reuse in the L1 cache. This difference leads to the selection of weight-stationary scheduling for x86, and input-stationary scheduling for POWER10.

\subsection{\name~Outperforms \base~in Every Model}
\label{sec:speedup}

\rtab{speedup} shows the convolution-only speedup and the whole-model speedup including every convolution from each model. 
\name~outperforms \base~in every model on both architectures with convolution speedups ranging from 12\% to 46\%.
As expected, higher speedups are observed in models with large convolutions, such as VGG-16 and ResNet-18. 
The fraction of the time spent in convolution in ONNX-MLIR determines the effect of using \name~over \base~in the whole-model performance. 
On average, whole models are between 15\% and 18\% faster. 

\begin{table}[!t]
\centering
\resizebox{\columnwidth}{!}{%
\begin{tabular}{|l|rc|rc|rc|}
\cline{1-7}
\multicolumn{1}{|c|}{\multirow{2}{*}{\textbf{Model}}} &
  \multicolumn{2}{c|}{\textbf{\begin{tabular}[c]{@{}c@{}}Convolution\\ Speedup\end{tabular}}} &
  \multicolumn{2}{c|}{\textbf{\begin{tabular}[c]{@{}c@{}}Model\\ Speedup\end{tabular}}} &
  \multicolumn{2}{c|}{\textbf{\begin{tabular}[c]{@{}c@{}}Convolution\\ Time Share\end{tabular}}}  \\ 
\multicolumn{1}{|c|}{} &
  \multicolumn{1}{c}{\textbf{x86}} &
  \multicolumn{1}{c|}{\textbf{P10}} &
  \multicolumn{1}{c}{\textbf{x86}} &
  \multicolumn{1}{c|}{\textbf{P10}} &
  \multicolumn{1}{c}{\textbf{x86}} &
  \multicolumn{1}{c|}{\textbf{P10}} \\ \cline{1-7}
GoogleNet \cite{googlenet} & \multicolumn{1}{r|}{1.18} & 1.26 & \multicolumn{1}{r|}{1.10} & 1.10 & \multicolumn{1}{r|}{0.49} & 0.43  \\ 
InceptionV2 \cite{inceptionv2} & \multicolumn{1}{r|}{1.21} & 1.28 & \multicolumn{1}{r|}{1.16} & 1.17 & \multicolumn{1}{r|}{0.76} & 0.62  \\ 
ResNet-18 \cite{resnet}  & \multicolumn{1}{r|}{1.26} & 1.46 & \multicolumn{1}{r|}{1.25} & 1.42 & \multicolumn{1}{r|}{0.93} & 0.82  \\ 
ResNet-50 \cite{resnet} & \multicolumn{1}{r|}{1.13} & 1.28 & \multicolumn{1}{r|}{1.09} & 1.17 & \multicolumn{1}{r|}{0.87} & 0.72  \\ 
ResNet-152 \cite{resnet} & \multicolumn{1}{r|}{1.16} & 1.28 & \multicolumn{1}{r|}{1.13} & 1.21 & \multicolumn{1}{r|}{0.91} & 0.77  \\ 
SqueezeNet \cite{squeezenet} & \multicolumn{1}{r|}{1.12} & 1.27 & \multicolumn{1}{r|}{1.16} & 1.13 & \multicolumn{1}{r|}{0.74} & 0.55  \\ 
VGG-16 \cite{vgg16}   & \multicolumn{1}{r|}{1.27} & 1.28 & \multicolumn{1}{r|}{1.16} & 1.12 & \multicolumn{1}{r|}{0.70} & 0.39  \\ \cline{1-7}
Geo Mean & \multicolumn{1}{r|}{1.19} & 1.30 & \multicolumn{1}{r|}{1.15} & 1.18 & \multicolumn{1}{r|}{0.76} & 0.59  \\ \cline{1-7}
\end{tabular}}
\caption{\name~performance speedup over \base~in machine-learning models.}
\label{tab:speedup}
\end{table}

\begin{figure*}[ht]
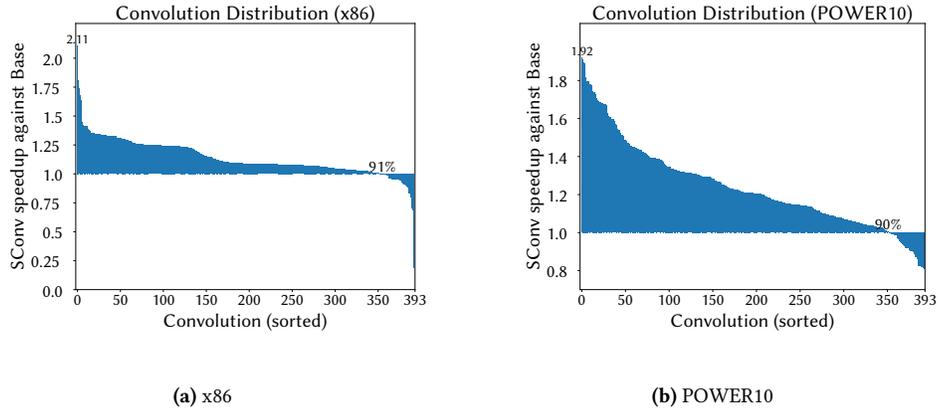

    \centering
    \subfloat[x86]{
        \scalebox{0.26}
        {\input{plotfiles/g4_x86_convolution_distribution.pgf}}
    }
    \subfloat[POWER10]{
        \scalebox{0.26}
        {\input{plotfiles/g4_power_convolution_distribution.pgf}}
    }
    \caption{\name~performance speedup against \base~for each of the 393 individual convolutions from the evaluated machine learning models. The x-axis is sorted by speedup (y-axis). \base's performance is normalized to 1. The graphs also indicate the percentage of convolutions in which \name~outperforms \base.}
    \label{fig:speedup-convs}
\end{figure*}

A more in-depth performance analysis needs to examine the performance changes in individual convolutions. 
\rfig{speedup-convs} shows the results for all individual convolutions sorted by speedup.
For about 90\% of the convolutions, \name~outperforms \base.
The maximum speedup for x86 is 2.1$\times$, and for POWER10 is 1.9$\times$.
\name~reaches 77\% of the theoretical peak throughput in POWER10 and 90\% in x86.
\rssec{limitations} explores what is causing \name~to underperform \base~in about 10\% of the convolutions.

\begin{figure*}[ht]
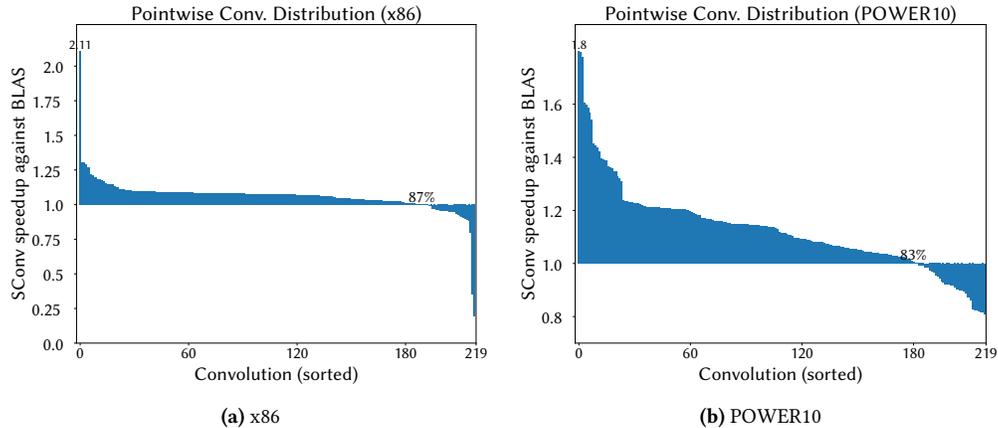

    \centering
    \subfloat[x86]{
        \scalebox{0.26}
        {\input{plotfiles/g9_x86_pointwise_distribution.pgf}}
    }
    \subfloat[POWER10]{
        \scalebox{0.26}
        {\input{plotfiles/g9_power_pointwise_distribution.pgf}}
    }
    \caption{\name~performance speedup against BLAS for each of the 219 pointwise convolutions from the evaluated machine learning models. The x-axis is sorted by speedup (y-axis). BLAS' performance is normalized to 1. The graphs also indicate the percentage of pointwise convolutions in which \name~outperforms BLAS.}
    \label{fig:pointwise-convs}
\end{figure*}



\subsection{Faster Micro-Kernels Make Packing Time More Relevant}
\label{sec:breakdown}

\begin{figure*}[ht]
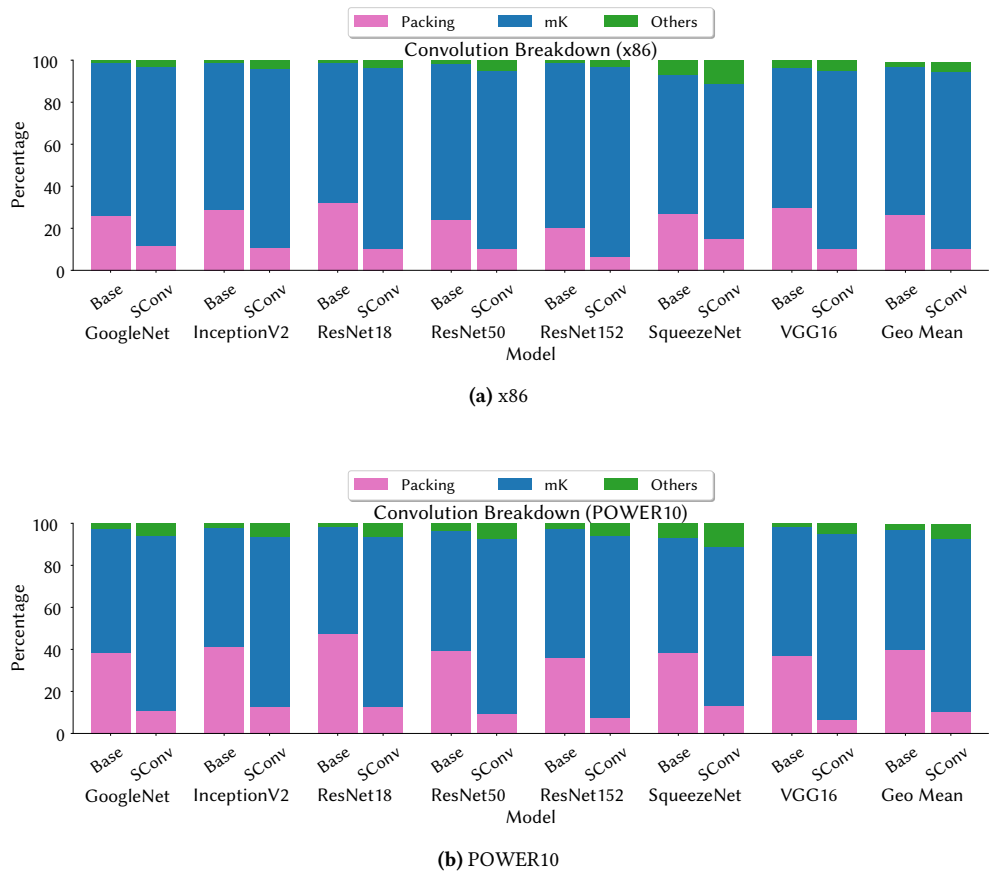

    \subfloat[x86]{
        \centering
        \scalebox{0.26}{\input{plotfiles/g3_x86_breakdown.pgf}}
    }
    \\
    \subfloat[POWER10]{
        \centering
        \scalebox{0.26}{\input{plotfiles/g3_power_breakdown.pgf}}
    }
    \caption{Breakdown of each model's total convolution time distributed between packing, micro-kernel (mK) and other steps (Others), for \base~and \name.}
    \label{fig:breakdown}
\end{figure*}

\rfig{breakdown} shows the convolution execution time divided into
\begin{inparaenum}[(a)]
\item "Packing", including Im2Col and GEMM packing;
\item "mK", the micro-kernel execution, which is the same for both approaches;
\item "Others", which represents the remaining boilerplate code (\tit{e.g.}, bias, control code, padding, etc.) that is different for \base~and \name.
\end{inparaenum}
In \name, packing has a smaller contribution to the execution time in comparison with \base. \name~provides three major improvements in packing: elimination of the Im2Col step, reduction of unnecessary loads by Vector-Based Packing, and optimization of filter-packing at compile-time.

The x86 results show that \base~has, on average, 26\% of its execution time dedicated to packing (20\% --- 32\%), while \name's packing share averages 10\% (6\% --- 15\%). 
The POWER10 results indicate a larger improvement, with \base's packing share averaging 39\% (36\% --- 47\%) and \name's averaging 10\% (7\% --- 13\%), which is a packing time improvement of 3.0x on x86 and 5.0x on POWER10.  The improvement range is 2.0x --- 3.9x on x86 and 3.6x --- 7.2x on POWER10.
Faster hardware-supported micro-kernel makes the packing time more significant in the overall execution. This effect is evident in the better POWER10 results, which translates to overall speedup in \rtab{speedup} and \rfig{speedup-convs}.

Packing data is most significant for large convolutions, such as those in ResNet18 and VGG16,  that contribute most to the model time. Large convolutions are the most sensitive to some of the improvements in \name, such as VBP and the elimination of Im2Col. Thus, packing optimizations are the main contributors to the results shown in \rssec{speedup}.

\subsection{Im2Col is Not Needed for Pointwise Convolutions}
\label{sec:limitations}

\tit{A pointwise convolution} comprises $1 \times 1$ windows with unitary stride. Such convolutions are frequently found in ML models --- $219$ out of the $393$ (55\%) convolutions in our experimental evaluation. Even though they are usually smaller and faster to compute, they often do not contribute as much to the model execution time. For a pointwise convolution, the Im2Col transformation produces a simple copy of the input matrix, and thus  Im2Col can be skipped altogether. 
A pointwise convolution can be trivially reduced to a GEMM operation, thus \name~is compared directly with the BLAS library's GEMM routine. Therefore, \name~does not have the advantage of fewer packing steps or Vector-Based Packing in these convolutions.

\name's results for these convolutions, shown in \rfig{pointwise-convs}, have a pattern that matches, to some degree, the one in \rfig{speedup-convs}, indicating good performance even when there is no Im2Col overhead in the baseline. 
\name~outperforms BLAS GEMM in 87\% of all pointwise convolutions on x86, and 83\% on POWER10, with larger average improvements in the positive cases than slowdowns in the negative ones. These results evidence the efficiency of CSA's tiling strategy for different convolution shapes and the advantage of packing filters at compile-time. Compile-time packing is only important in convolutions with many filters and channels and small $\mathit{IN}_h$ and $\mathit{IN}_w$, and this is reflected in which convolutions achieve higher speedups.


Most convolutions for which \base~outperforms \name~are pointwise: 93\% on POWER10 and 78\% on x86.
For instance, the largest pointwise convolutions found in the tested models have $\mathit{IN}_h = \mathit{IN}_w = 56$, and are consistently faster in \base. These cases fit the best-performance scenario of the BLAS GEMM routine because the inner dimension ($\mathit{IN}_c$) is small compared to the dimensions of the output matrix \cite{direct_conv}. Furthermore, filter packing has a smaller impact on the overall execution time when the input tensor is large.

\subsection{\name~Reduces Cache Misses in All Levels of Cache}
\label{sec:locality}



\begin{figure*}[ht]
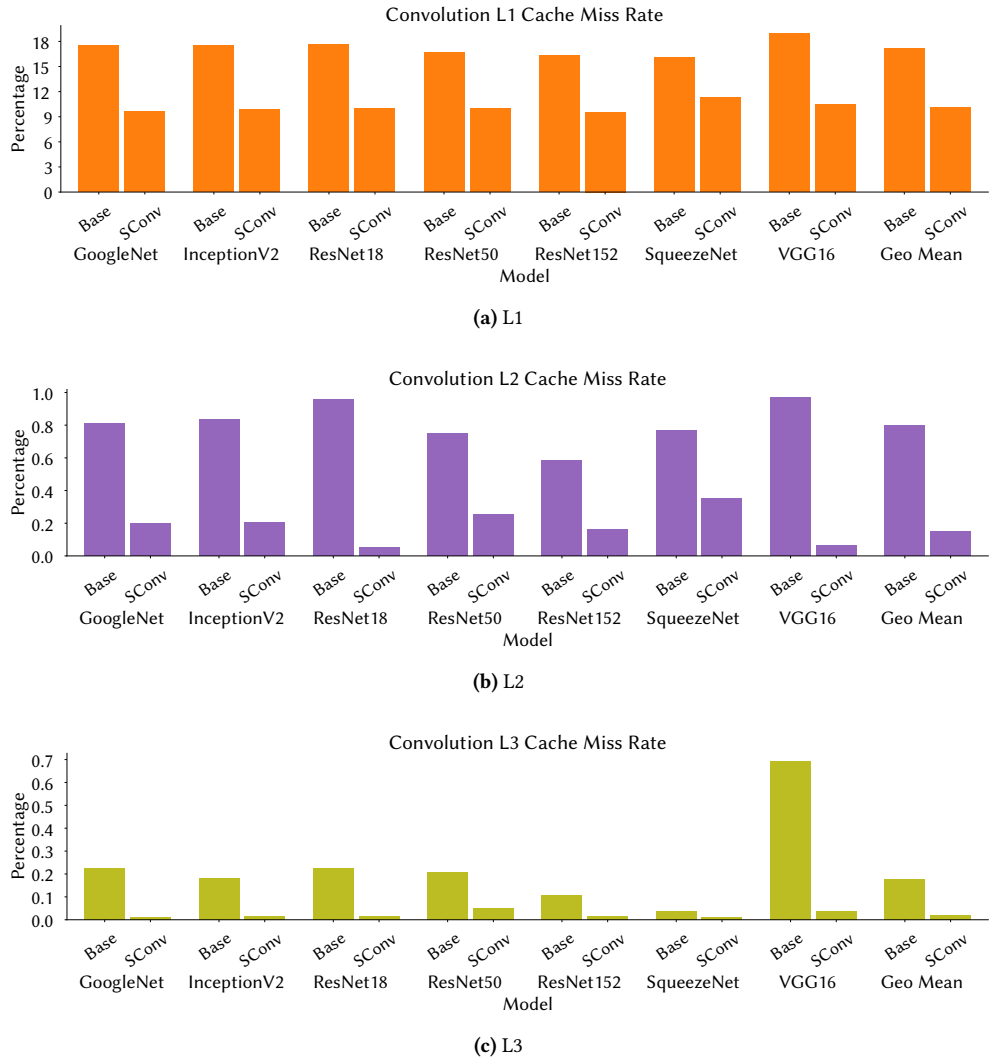

    \subfloat[L1]{
        \centering
        \scalebox{0.26}{\input{plotfiles/g10_x86_l1_cache_miss_breakdown.pgf}}
    }
    \\
    \subfloat[L2]{
        \centering
        \scalebox{0.26}{\input{plotfiles/g10_x86_l2_cache_miss_breakdown.pgf}}
    }
    \\
    \subfloat[L3]{
        \centering
        \scalebox{0.26}{\input{plotfiles/g10_x86_l3_cache_miss_breakdown.pgf}}
    }
    \caption{Breakdown of each model's cache miss performance in each level, for \base~and \name~on x86. Shows the percentage of cache misses in each level from the total loads.}
    \label{fig:cache-breakdown}
\end{figure*}

The main goal of the slicing strategy in \name~is to improve cache performance. 
The Im2Col step is bad for cache performance since most tiles, after expansion, need to be reloaded from a more distant cache level or DRAM for GEMM packing and computation. Removing Im2Col allows for Packing On Demand, which improves cache locality by keeping the tiles in the cache after packing for the micro-kernel. Furthermore, the tiling strategy determined by CSA optimizes tile reuse by scheduling in order to minimize the data flow between memory levels. However, the addition of explicit padding in \name~adds extra loads which may lead to more cache misses.

\name~has fewer loads than \base~in 65\% of all tested convolutions (246). Even when \name~has more loads than \base~(147, 35\%), the impact on cache misses is not significant: \name~has more L1 misses on 0.5\% of the convolutions (2), more L2 misses on 5\% (20) and more L3 misses on 4\% (15). In these cases, the difference is small.

The miss rates shown in~\rfig{cache-breakdown} are for all loads in the model. 
For \name, around 90\% of all loads resolve in L1 cache, compared with 83\% for \base~--- a 1.9$\times$ improvement.
The difference between the methods increases at lower cache levels. 
On average, \name~has 5.9x fewer L2 misses and 9.9x fewer L3 misses, though both represent a small percentage of the total number of loads. 
In the case of VGG-16, \name's improvements on larger convolutions resulted in a big drop in L3 misses because, with the input not fitting in cache, data has to be reloaded from DRAM in \base, which does not happen in \name.
Not only \name~is better at minimizing expensive memory operations, it also often requires fewer load operations to compute the convolution.

\section{Related Work}
\label{sec:related-work}

 Chellapilla \etal proposed the computation of convolution using the Im2Col transformation to reduce the problem to GEMM and solve it using a BLAS library~\cite{im2col}. 
 This approach is used in many machine learning frameworks, such as TensorFlow \cite{tensorflow}, PyTorch~\cite{pytorch} and Caffe~\cite{caffe}. 
 The BLAS's GEMM routine already contains packing steps, thus two separate steps require data manipulation: expansion (Im2Col) and packing. 
 This approach has a large memory footprint because of the large matrix resulting from Im2Col.
 Also, the GEMM tiling can be inefficient when the inner dimension is large. 
 Other authors address the memory footprint issue while still using BLAS GEMM as a foundation to solve the convolution problem \cite{kn2row, mec, kn2row2}. 
 However, they still use GEMM's tiling with multiple data-manipulation steps.

Zhang \etal\cite{direct_conv} discuss the limitations and inefficiencies of the Im2Col + BLAS convolution algorithm and present an argument for using direct convolution instead. 
They use a model architecture with fused multiply-add instructions to explore cache blocking, and data reuse, given a friendly memory layout. 
Unlike the work proposed by Zhang \etal, \name~explores more scheduling strategies and tile distribution in the cache hierarchy, provides a template that can be used in different architectures and does not impose a different memory layout.
Thus \name~can be deployed in other AI/ML frameworks without changing any other operator implementation.

Goto and de Gejin~\cite{goto} introduce the idea of exploiting an optimized micro-kernel (called "inner-kernel") by tiling the problem in an external macro-kernel. These tiles are then packed to a friendly layout for the micro-kernel. 
The CSA algorithm resembles the algorithm proposed by Goto and de Gejin in its functionalities, but it applies the strategy to convolution instead of applying it to GEMM.

Juan \etal\cite{convgemm} modify the BLIS library's~\cite{blis} GEMM routine to apply Im2Col in its packing step. The result is a convolution and GEMM hybrid named ConvGEMM, which takes advantage of Goto and de Gejin's GEMM tiling and structure with the Im2Col transformation applied on the fly, eliminating the double-packing problem. This process is similar to Packing on Demand, but the tiling is still GEMM-based, that is, 2D slicing without the concept of scheduling, and therefore has different cache usage, tile reuse, and packing patterns. A similar attempt to adapt GEMM to a convolution algorithm by lazy Im2Col packing was made for GPU hardware by Chetlur \etal \cite{cudnn}. Dukhan  proposes an indirect-convolution algorithm that modifies GEMM all the way to the micro-kernel to avoid materializing the Im2Col matrix in memory~\cite{indirect}. 

Korostelev \etal\cite{yaconv} combine the ideas of modifying the GEMM routine and decomposing convolution into multiple GEMM operations~\cite{kn2row2} to create a new method that avoids packing redundancy while keeping the overall GEMM structure (tiling, packing, micro-kernel). Contrary to the approach in \name, \cite{yaconv} works only on unitary stride convolutions and improves only non-pointwise convolutions. Moreover, their experiments focus on isolated convolutions and do not provide results for complete models.

Space exploration tools such as TVM~\cite{tvm} and Ansor~\cite{ansor} can change loop order and tile size using machine-learning techniques and a cost model. Due to the highly automated nature of these tools, total hardware usage is harder to achieve in more complex architectures. The ample tiling space also leads to long compilation and code-generation times, hindering the ML model design-time efficiency.

This project follows the work done by Sousa \etal\cite{csa_nmp} in tiling 3D convolutions targeting NPUs, by generalizing its tiling analysis for CPUs in the CSA pass. In that work, the Input Stationary and Weight Stationary scheduling strategies are used in a different context, and the modeling focuses on taking advantage of memory bursts while loading data to the scratchpad memories.

Following the work from Zhang \etal\cite{direct_conv}, Barrachina \etal\cite{direct_arm}  propose two new direct-convolution algorithms for the NHWC layout (batch $N$, height $H$, width $W$, and channels $C$)  on ARM processors. 
Like \name, they tile in the channel dimension and use a BLAS micro-kernel. 
In \name, however, besides considering the partitioning of the input tensor and of the filter set, tiling also considers the two scheduling alternatives --- input or weight stationary --- and the tile distribution over the cache hierarchy. 
\name~shows that direct convolution can benefit from:
\begin{inparaenum}[ (a)]
\item novel ISA extensions, such as the POWER10 MMA, for faster tensor-algebra computation;
\item vector-register extensions, such as Intel x86 AVX and POWER VSX, to implement Vector-Based Packing.
\end{inparaenum}
The experimental evaluation demonstrates the flexibility of \name~by using two different CPU architectures and the common NCHW layout.

\section{Conclusion and Future Work}
\label{sec:conclusion}
As stated in the introduction, the objective of the algorithm presented in this paper is to provide a good implementation of direct convolution for different CPU architectures. \name~achieves this goal, outperforming Im2Col + BLAS in experiments performed in realistic conditions with full machine learning model inference (\rssec{speedup}) on both tested architectures by reducing total data manipulation time (\rssec{breakdown}). \name~also outperforms regular BLAS GEMM when computing pointwise convolutions in over 83\% of the 219 instances tested (\rssec{limitations}).

As for future work, some interesting paths can be pursued. Currently, the CSA tiling analysis pass creates tiles based solely on micro-kernel information and the $\mathit{IN}_c$ value. More robust space exploration can be done by exploiting other dimensions of $\mathit{IN}^T$ and $\mathit{FS}^T$. This change may further improve cache locality by better utilizing the L1 cache. CSA's optimization heuristic may also be improved if a binary search is used to find a better approximation for the tile size. Moreover, since the approach is integrated with machine-learning compilers, a fully code-generation-based implementation can leverage compile-time information to simplify operations such as padding and remove boilerplate code overhead. Finally, a hybrid solution with Im2Col + BLAS might perform best for each convolution.

\bibliographystyle{ACM-Reference-Format}
\bibliography{sconv}

\end{document}